\documentclass[runningheads,a4paper]{llncs}
\usepackage{wrapfig, framed, caption}
\usepackage{amssymb}
\usepackage{amsmath}
\usepackage{graphicx}
\usepackage{epstopdf,multirow}
\usepackage{array}
\usepackage{cite,url}
\usepackage{marvosym}

\usepackage{url}
\urldef{\mailsa}\path|{alfred.hofmann, ursula.barth, ingrid.haas, frank.holzwarth,|
\urldef{\mailsb}\path|anna.kramer, leonie.kunz, christine.reiss, nicole.sator,|
\urldef{\mailsc}\path|erika.siebert-cole, peter.strasser, lncs}@springer.com|    
\newcommand{\keywords}[1]{\par\addvspace\baselineskip
\noindent\keywordname\enspace\ignorespaces#1}

\begin{document}
\pagestyle{empty}
\mainmatter

\title{CapsDeMM: Capsule network for Detection of Munro\textquoteright s Microabscess in skin biopsy images}

\author{Anabik Pal$^{1(\text{\Letter})}$
\and Akshay Chaturvedi$^{1}$ \and Utpal Garain$^{1}$\and  Aditi Chandra$^{2}$\and Raghunath Chatterjee$^{2}$\and
 Swapan Senapati$^{3}$}

\institute{$^{1}$CVPR Unit, Indian Statistical Unit, Kolkata 700108, West Bengal, India\\
\textbf{anabikpal@gmail.com} \\
$^{2}$Human Genetics Unit, Indian Statistical Unit, Kolkata 700108, West Bengal, India\\
$^{3}$Consultant Dermatologist, Uttarpara, Hooghly, 712258, West Bengal, India\\
}


\maketitle

\begin{abstract}

This paper presents an approach for automatic detection of Munro\textquoteright s Microabscess in stratum corneum (SC) of human skin biopsy in order to realize a machine assisted diagnosis of Psoriasis. The challenge of detecting neutrophils in presence of nucleated cells is solved using the recent advances of deep learning algorithms. Separation of SC layer, extraction of patches from the layer followed by classification of patches with respect to presence or absence of neutrophils form the basis of the overall approach which is effected through an integration of a U-Net based segmentation network and a capsule network for classification. The novel design of the present capsule net leads to a drastic reduction in number of parameters without any noticeable compromise in the overall performance. The research further addresses the challenge of dealing with Mega-pixel images (in 10X) vis-à-vis Giga-pixel ones (in 40X). The promising result coming out of an experiment on a dataset consisting of $273$ real life images shows that a practical system is possible based on the present research. The implementation of our system is available at~\url{https://github.com/Anabik/CapsDeMM}.

\end{abstract}

\section{Introduction}\label{Sect::Introduction}

Psoriasis is a chronic, immune-mediated, relapsing, inflammatory skin disease with variable morphology, distribution, severity and course~\cite{Marks2012}. The prevalence of psoriasis varies 1\%-12\% among different populations worldwide. It is often difficult to differentiate psoriasis from other erythemato-squamous diseases like Seborrheic dermatitis, Leprosy, Lichen planus, Tinea corporis, Pityriasis, Eczema etc.~\cite{Anabik2016,Roy2017,Pal2018}. Hence, histopathological examination is considered for confirmation.

Psoriasis develops when the immune system mistakes a normal skin cell for a pathogen and sends out faulty signals to yield the over production of new skin cells. Hence, in many cases, neutrophils and nucleated cells infiltrate into stratum corneum (SC). This infiltration occurs either in confluent (throughout the SC layer) or in focal (not confluent) manner. The presence of nucleated cells in SC is termed as parakeratosis and the accumulation of neutrophils in SC along with parakeratosis is termed as Munro\textquoteright s Microabscess (MM). In clinical pathology, Munro\textquoteright s Microabses is considered as the diagnostic hallmark of psoriasis~\cite{Marks2012}.

\begin{figure}[htp]
\centering
\fbox{\includegraphics[height=.8in]{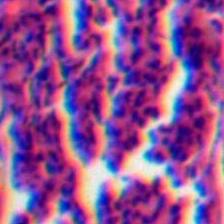}}\quad \fbox{\includegraphics[height=.8in]{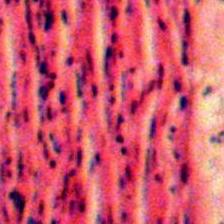}}\quad \fbox{\includegraphics[height=.8in]{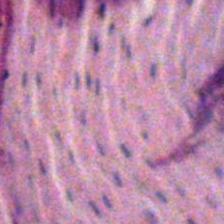}}\quad \fbox{\includegraphics[height=.8in]{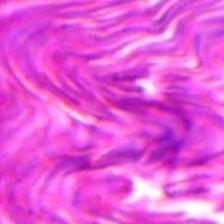}}\\
\hspace{0in}(a) \hspace{.8in} (b)  \hspace{.8in} (c)  \hspace{.8in}(d) \\

\caption{Patches cropped from stratum corneum layer of WSIs. Neutrophils are circular shaped and dark blue colored. Nucleus are oval shaped and light blue colored. Here, in (a) and (b) both neutrophils and nucleated cells are present, in (c) only nucleated cells are present and in (d) neither  nucleated cells nor neutrophils are present.}
\label{fig:Neutrophil Images}
\end{figure}

Challenge in detecting MM lies in the fact that due to staining variation, neutrophils in stratum corneum are often misclassified as nucleated keratinocytes since both of them become dark stained. Several imaging artefacts add further challenges and thus, accurate diagnosis of skin biopsy by eye-inspection is challenging, even for highly experienced pathologists. Fig~\ref{fig:Neutrophil Images} illustrates the problem.

\vspace{-.5cm} 

In the last two decades, several automatic systems are designed and developed to complement the workload of pathologists from microscopic examination of clinical tissue. But there is a dearth in automatic identification of neutrophils in biopsy images. Some research initiatives are reported for acute inflammation diagnosis~\cite{Wang2014,Wang2015} where Giga-pixel images are used.

The contributions of this paper are many folds. The pathological challenges of detection of MM in skin biopsy have been solved by developing an automated computational framework incorporating the latest advances in deep learning. The capsule network has been designed in such a way that drastically reduces the number of parameters without sacrificing performance. Mega-pixel images are used instead of usual Giga-pixel ones to reduce the computational burden and thereby supporting a low cost imaging system. Preparation of an annotated dataset of $273$ whole slide skin biopsy images (WSI) is another important outcome of this research. This dataset\footnote{For the dataset contact Prof. Utpal Garain (utpal@isical.ac.in)} not only helps to demonstrate the efficiency of the present approach, but would also facilitate further research.

\section{Proposed Methodology}\label{Sect::Proposed Methodology}
\begin{figure}[!h]
\center
{\includegraphics[width=0.8\textwidth]{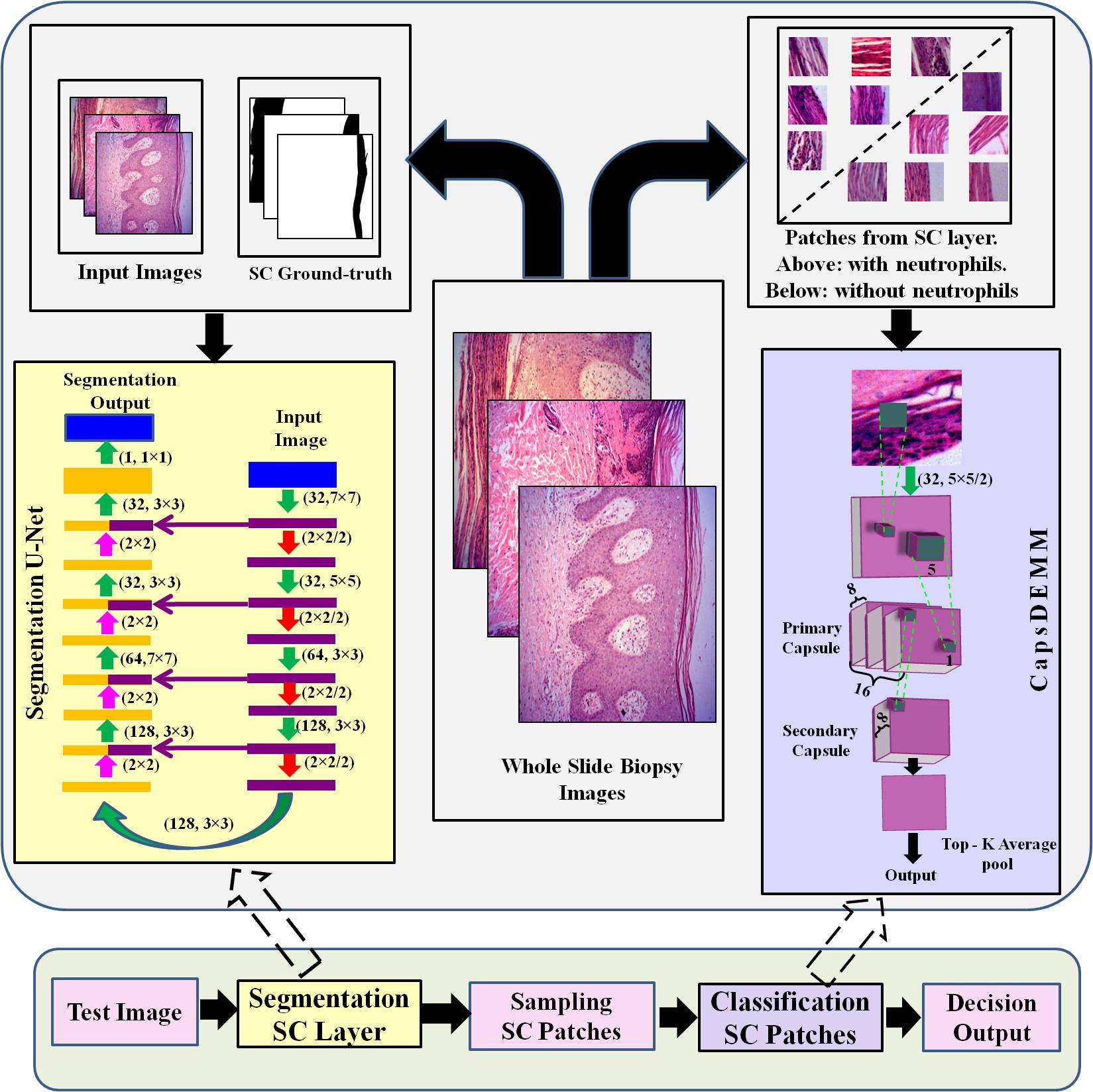}}
\caption{Proposed system architecture. The green arrows represent convolution layer followed by ReLU activation, red arrows represent max-pooling layer, pink arrows represent up-sampling layer and dark purple arrow represent skip connection.}
\label{fig:System Architecture}
\end{figure}
The goal of this paper is to detect Munro\textquoteright s Microabscess in whole slide skin biopsy images (WSI). We break down the task into three parts (i) Segmentation of SC layer, (ii) Patch extraction from SC layer and (iii) Neutrophil detection in the extracted patches. Note that we are only interested in detecting neutrophils present in the SC layer as opposed to the entire image and the SC layer lies in a small portion of the WSI image (approx. 2-15\%). Hence, segmentation of SC layer followed by neutrophil detection is a logical approach. The pictorial representation of the proposed framework is shown in Figure~\ref{fig:System Architecture}. A brief description about  proposed framework is given in the following subsections.

\subsubsection{Stratum corneum segmentation}\label{section:: Stratum sorneum segmentation}
Nowadays, U-Net~\cite{Ronneberger2015} is the state-of the art for image segmentation. We trained the U-Net given in Figure~\ref{fig:System Architecture} for segmenting the SC layer. Given the segmentation output $S_{x,y}$ and the corresponding ground truth $G_{x,y}$, we minimize the dice loss function $L_{seg}$ given by

\begin{equation}
L_{seg} = \frac{2\times\sum_{x,y} G_{x,y}S_{x,y}}{\sum_{x,y} G_{x,y} + \sum_{x,y} S_{x,y}}
\end{equation}

where $x, y$ denote the spatial coordinate of the WSI image.

\subsubsection{Stratum corneum patch selection}\label{section:: Stratum sorneum patch selection}
The stratum corneum patches should be selected in such a way that all the pixels of a patch belong to a perceptually similar region as well as their union covers the entire stratum corneum (SC). For dividing the image into perceptual regions, Simple Linear Iterative Clustering (SLIC)~\cite{Achanta2012} super-pixel algorithm is applied. Then a square window  around the centroids of the super-pixels which lie in the stratum corneum are selected.

\subsubsection{CapsDeMM for Stratum corneum  patch classification}\label{section:: CapsDeMM for Stratum corneum patch classification}
Recently, convolutional neural networks have achieved state of the art for several image classification tasks. But the max pooling operation used in traditional convolutional neural network architecture may ignore important spatial information cues which is undesirable.
So, in this paper, recently introduced capsule network~\cite{Sabour2017} is adopted for SC patch classification. Capsule network uses \textquotedblleft routing by agreement\textquotedblright policy to ensure that significant spatial information in an image is not lost as we go from lower to higher layers.

The capsule network consists of two parts, namely, primary capsule and secondary capsule. We designed the capsule network in such a way that the receptive fields of the capsules in the secondary capsule avoid crowding. Crowding in capsule networks refers to the phenomenon where multiple instances of the same entity is present in the receptive field of a capsule. In such a case, the capsule is unable to encode the instantiation parameters of the concerned entity. The length of the output vector of secondary capsule denotes the probability of neutrophil in a particular portion of the image patch. There can be several neutrophils in a particular image patch. Keeping this in mind, average of the top $K$ probabilities (top-K average pooling) is considered as the probability of neutrophil in the given image patch. Given an image patch $I$, let $p_{I}$ denote the probability of neutrophil in the image patch.  Then, during training, we minimize the binary cross entropy loss function $L_{patch}$ given by

\begin{equation}
L_{patch} = - y_{I}* log(p_{I}) - (1 - y_{I})* log(1-p_{I})
\end{equation} 

where $y_{I}$ is $1$ if the image patch contains neutrophil and $0$ otherwise. This network is further referred to as CapsDeMM (\textbf{Caps}ule network for \textbf{De}tection of \textbf{M}unro\textquoteright s \textbf{M}icroabcess).

\section{Dataset}\label{Sect::Dataset}
In this research, after clinical confirmation of psoriasis, affected tissues are collected in 10\% formalin under the supervision of an expert dermatologist. Formalin fixed tissues are dehydrated and embedded in paraffin blocks. Thin sections (5 $\mu$M) are used for slide preparation and then stained with hematoxylin and eosin to prepare the histopathological slides. These slides are kept
under a microscope with 10X magnification and the images are collected from the microscope using a digital camera. The images are captured in 10X magnification as this is the highest magnification in which the whole biopsy sample fits adequately in the field of view of the camera. The size of the captured images are $1936 \times 2584$ pixels. Written informed consents were obtained from all patients before recruiting them for the study. This study is conducted after obtaining ethical approval from the ‘Review Committee for Protection of Research Risks to Humans’ of Indian Statistical Institute, Kolkata, West Bengal, India.

The skin biopsy samples are collected from $120$ patients. Multiple serial sections of skin tissue present in the biopsy slides are imaged. The images where the stratum corneum (SC) layer is lost during tissue processing are discarded. Then the images are labelled by two experts. The images where both experts' agreement match are considered for this research. In our dataset, there are $88$ images with Munro\textquoteright s Microabscess and $185$ images without Munro\textquoteright s Microabscess. The ground-truth annotation of SC segmentation of the images is done by an expert. In order to construct the SC patch classifier, multiple squared patches ($224\times224$ pixel sized) are cropped from SC layer of the biopsy images. Then the existence of neutrophil in these patches are labelled by two experts and the cases where both experts' agreement match are chosen. In total, there are $886$ patches with neutrophils and $1700$ patches without any neutrophil. In rest of the paper, the SC patches having neutrophils are termed as positive patches and the patches not having neutrophils are termed as negative patches.

\section{Experiments}\label{Sect::Experiments}
\subsection{Experimental Setting}\label{Sect::Experimental Design}
The proposed system is tested with three-fold cross validation. Each fold contains random selection of $91$ images from our dataset. Among them, two folds contain $29$ images and another fold contains $30$ images having Munro\textquoteright s Microabscess. The cropped SC patches are grouped fold-wise ($862$ patches/fold) to build the fold-wise SC patch classifiers. The validation data is developed by random selection of 10\% training images (original WSI) for U-Net and 20\% training images (SC patches) for CapsDeMM. The validation data is used for tuning filter number, filter size and other hyper-parameters.  The architecture of the used U-Net and capsule network are shown in Figure~\ref{fig:System Architecture}. In CapsDeMM, each primary capsule contains 8 convolutional units ($5\times5$ and stride 2) of 16 channels and the secondary capsule is a convolutional capsule which does routing by agreement between the capsules in the same spatial region of the primary capsule. For U-Net, keeping the resource constraints in mind, the original images and the corresponding ground-truths are down-sampled to $960\times1280$. However, the segmentation output is up-scaled to the original dimension and $224\times224$ pixel sized SC patches are selected for classification. In order to achieve the best performing CapsDeMM network,  the value of $K$ for the top-K average pooling layer is tuned (without making any architectural change to the other layers) and the resulting network is named as CapsDeMM-K.

\vspace{.5cm}

\subsection{Results and Discussion}\label{section::Result and Discussion}

\begin{figure}[!h]
\centering
\fbox{\includegraphics[height=1.25in,width=1.75in]{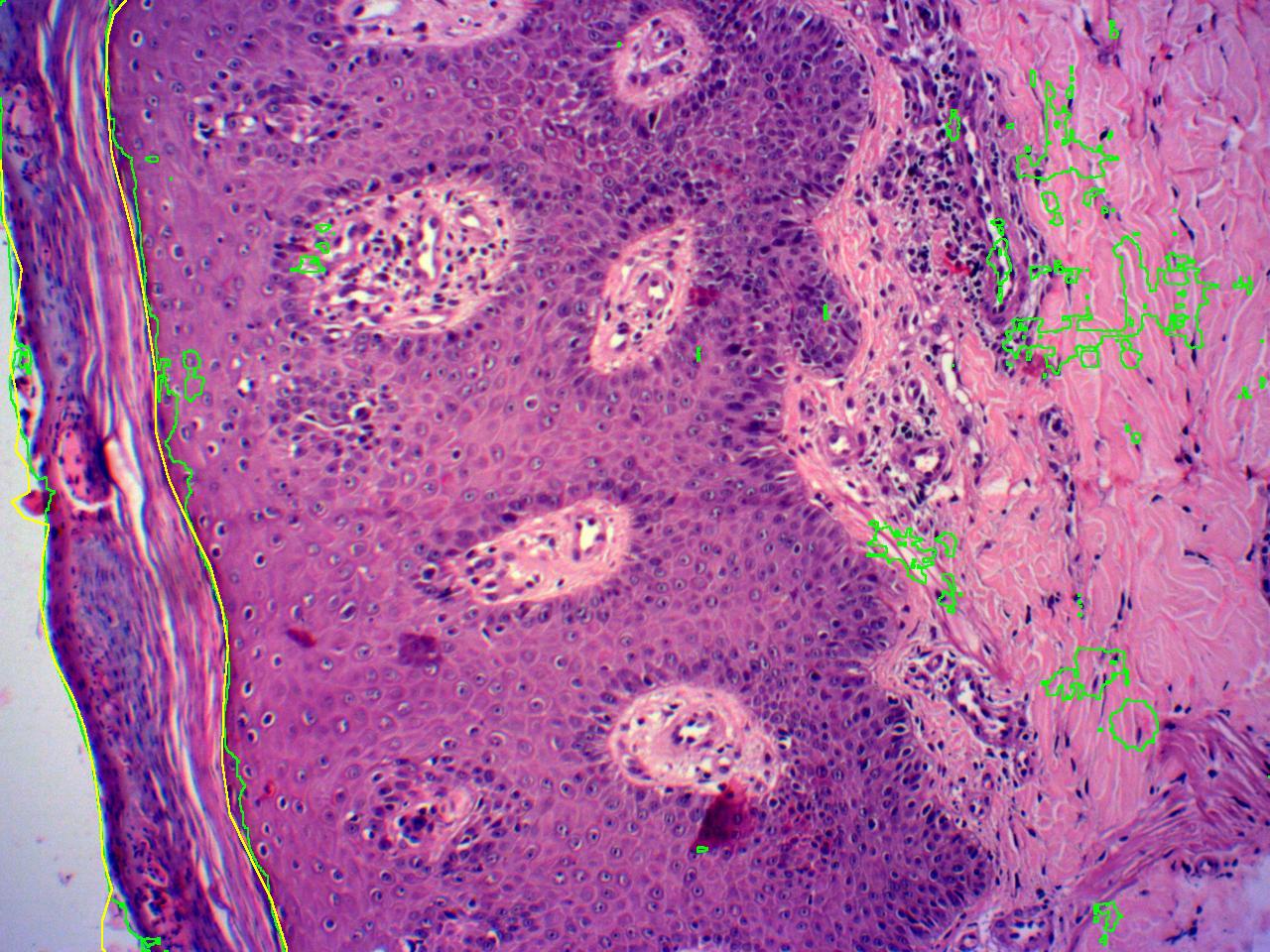}}\quad \fbox{\includegraphics[height=1.25in,width=1.75in]{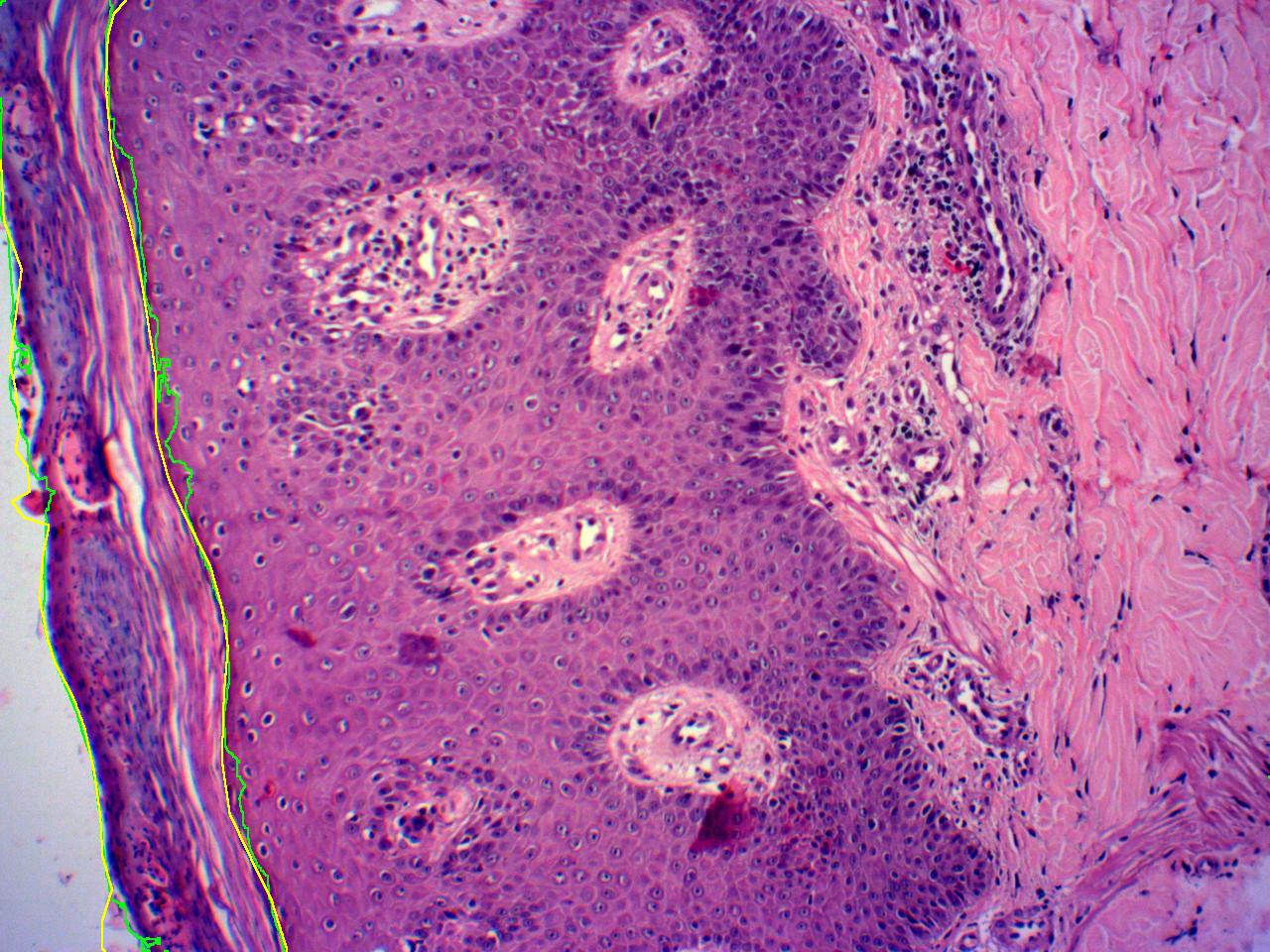}}\\
{(a)} \hspace{2in}  (b)
\caption{The yellow line is used to denote the ground-truthed region boundary whereas the green line is used to denote the detected region boundary. (a) Segmentation outcome from the U-Net, (b) Segmentation outcome after post-processing}
\label{fig:Segmentation Images}
\end{figure}

\paragraph{\textbf{Segmentation Performance:}}\label{section:: Segmentation Performance}
The accurate segmentation of the SC region improves the diagnosis performance of our system. The U-Net produces good segmentation (Dice Coefficient of 0.8493$\pm$0.025) but it generates several spurious holes and isolated segmented regions in the SC regions as shown in Figure~\ref{fig:Segmentation Images}. To get rid of it, \textquoteleft hole-filling' algorithm is applied for smoothing such spurious and isolated regions. Figure~\ref{fig:Segmentation Images} illustrates that the used smoothing technique is able to remove falsely detected non-SC regions. Final segmentation outcome results in Dice Coefficient of 0.8614$\pm$0.014.

\begin{minipage}{\textwidth}
  \begin{minipage}[b]{0.38\textwidth}
    \centering    
    \fbox{\includegraphics[height=1.3in,width=1.6in]{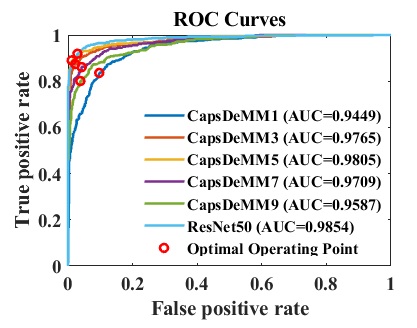}}
    \captionof{figure}{Comaprison of ROC curves.}\label{fig:ROC}
  \end{minipage}
  \begin{minipage}[b]{0.58\textwidth}
    \centering
    {\begin{tabular}{| >{\centering\arraybackslash}m{1in} | >{\centering\arraybackslash}m{.77in} | >{\centering\arraybackslash}m{.77in}|}    
		\hline
		Metric&	ResNet-50& 			CapsDeMM5\\
		\hline
		Recall&	$0.87\pm 0.078$& $0.84\pm 0.064$\\
		\hline
		Precision&$0.95\pm0.015$&$0.94\pm0.027$\\
		\hline
		F1 Score&$0.91\pm 0.036$&$0.88\pm 0.027$\\
		\hline	
		Accuracy&$0.94\pm 0.020$&	$0.92\pm 0.013$\\
		\hline	
		Parameters&	23.5 M&	0.1 M\\
		\hline
		Load Time&2.15 s	&0.26 s	\\
		\hline
		Prediction Time&0.021 s/img	&0.087 s/img	\\
		\hline
	\end{tabular}  
	\captionof{table}{Performance comparison of patch classifiers. M = Million, s = second.} \label{table::Performance patch classifiers}  } 
    \end{minipage}
  \end{minipage}

\paragraph{\textbf{Stratum corneum patch classification:}}\label{section:: Caps Net for Stratum corneum patch classifier development}
The development of stratum corneum patch classification is an important component for the success of the proposed system. The capsule network shown in  Figure~\ref{fig:System Architecture} is used for this purpose. In $top-K$ average pooling, lower values of $K$ might misclassify an image patch as positive due to some portions of the patches getting high probabilities whereas higher values of $K$ might overcompensate this effect leading to positive samples being classified as negative. To get the optimum value of $K$, we compare the ROC curves for different values of $K$. $K=5$ is chosen for classification since it provides best AUC score (average of three folds). The diagnosis for presence or absence of neutrophils in a patch is made by comparing the cut off value obtained from the ROC analysis. In case the network output for a patch is less than the cut off value, the predicted output is negative, otherwise, the predicted output is positive. The comparison of ROC curves for first fold for five different $K$ values ($1$, $3$, $5$, $7$ and $9$) is shown in Figure~\ref{fig:ROC}.

The classification performance is evaluated with Recall, Precision, F1 Score and classification accuracy (ACC). The average value for all metrics for all three folds are listed in Table~\ref{table::Performance patch classifiers}. Finally, the performance of the proposed capsule network is compared with a trained CNN i.e. ResNet-50, trained on the same dataset. The experimental result shows that  Capsule network achieves comparable accuracy to ResNet-50 despite having orders of magnitude less parameters.

\paragraph{\textbf{Whole Slide Image (WSI) Diagnosis:}}\label{section:: WSI Classification}
Ideally, detection of a single positive SC patch should indicate the presence of Munro\textquoteright s microabscess. But there are occasional misclassification in classifying SC patches (see Table~\ref{table::Performance patch classifiers}). So, to develop a robust WSI Classification system, a threshold $T$ is decided from the training set. Only those WSIs which have more than $T$ number of positive patches are diagnosed as having Munro\textquoteright s microabscess. Two different strategies are analysed for selecting the value of $T$ - (I) The system should produce best WSI classification performance: $T$ is selected in such a way that the best WSI classification accuracy is achieved; (II) The system will reduce the workload of pathologists by rejecting the negative cases (slides without having Munro\textquoteright s Microabscess): $T$ is selected in such way  that true negative rate is maximized. Strategy I is evaluated with correct classification accuracy (ACC) and strategy II is evaluated with True Negative Rate (TNR) and Precision. Obviously, the value of $T$ varies across the folds depending on the used thresholding strategy.

In order to compare the performance of the proposed diagnostic system on the super-pixel numbers, $3$ different number of super-pixels are considered. The performance averaged over all the folds is listed in Table~\ref{Table:Performance comparison of WSI classifiers}. According to Table~\ref{Table:Performance comparison of WSI classifiers}, both capsule network and ResNet-50 produces comparable performances. Note that when Strategy I is considered, CapsDeMM outperforms ResNet-50 but when Strategy II is considered, ResNet-50 outperform CapsDeMM.

\begin{table}[]
\centering
\caption{Performance comparison of WSI classifiers}
\begin{tabular}{|c|c|c|c|c|c|c|}
\hline
\multirow{3}{*}{\begin{tabular}[c]{@{}c@{}}Number \\ of\\  Superpixel\end{tabular}} & \multicolumn{3}{c|}{ResNet-50}                & \multicolumn{3}{c|}{CapsDeMM5}                \\
\cline{2-7}
                                                                                    & Strategy I & \multicolumn{2}{|c|}{Strategy II} & Strategy I & \multicolumn{2}{c|}{Strategy II} \\
 \cline{2-7}                                                                                   & ACC (\%)   & TNR           & Precision       & ACC (\%)   & TNR           & Precision       \\
 \hline
300                                                                                           & 86.13      & 1.0000        & 0.4649       & 88.71      & 0.9606        & 0.4373   \\
 \hline
500                                                                                           & 85.76      & 0.9837        & 0.4479       & 88.70      & 0.9656        & 0.4291   \\
 \hline
700                                                                                           & 87.59      & 0.9651        & 0.5068   & 89.06      & 0.9333        & 0.4000\\
 \hline      
\end{tabular}\label{Table:Performance comparison of WSI classifiers}
\end{table}

\section{Conclusion and Future work}\label{sect::Discussion and conclusion}
This paper presents the first of its kind system for detection of Munro\textquoteright s Microabscess in skin biopsy images. The drastic reduction of parameters without notable performance degradation justifies the applicability of CapsDeMM for the present problem. The promising performance of the two strategies for WSI classification presented in the paper shows their applicability for reducing the workload of the pathologists by a huge margin. The use of Mega-pixel images not only reduces the overall computational burden but also attests the use of a low cost system consisting of a light microscope (without digital scanner) and a digital camera. The outcome of the present research along with the dataset of WSIs will help in addressing several other important histopathological analysis of psoriasis including classification of parakeratosis (confluent/focal), detection of Kogoj Microabscesses. Efficiency of the present framework can be validated by employing different architectures and the patch classification can also be attempted with patches of different shapes (e.g. rectangular).

\subsubsection*{Acknowledgments.} 
Authors would like to acknowledge all volunteers who participated in this study.

\bibliographystyle{plain}
\bibliography{references}{}

\end{document}